\documentclass[10pt,twocolumn,letterpaper]{article}

\usepackage[pagenumbers]{iccv}              %

\usepackage{comment}
\usepackage{lipsum}
\usepackage{tabularx} %
\usepackage{booktabs} %
\usepackage[labelsep=period,font=small]{caption} %
\usepackage{tikz} %
\usepackage{tikzfill} %
\usepackage{pgfplots} %
\usepackage{pgfplotstable} %
\usepackage{standalone} %
\usepackage{multirow,makecell} %

\newcolumntype{Y}{>{\centering\arraybackslash}X} %
\newcolumntype{P}[1]{>{\centering\arraybackslash}p{#1}} %
\def\checkmark{\tikz\fill[scale=0.4](0,.35) -- (.25,0) -- (1,.7) -- (.25,.15) -- cycle;} %
\def\crossmark{\tikz{\draw[line width=1pt](0,0.2) -- (0.2,0);\draw[line width=1pt](0,0) -- (0.2,0.2);}} %

\usetikzlibrary{spy} %
\usetikzlibrary{arrows.meta} %
\usepgfplotslibrary{polar} %
\usepgfplotslibrary{fillbetween}
\usetikzlibrary{patterns}
\usepgfplotslibrary{colorbrewer}
\usetikzlibrary{positioning}
\usetikzlibrary{shapes.geometric, arrows, spath3, intersections, calc}

\pgfplotsset{compat=newest} %

\definecolor{3787CF}{RGB}{55,135,207}
\definecolor{F1C232}{RGB}{241,194,50}
\definecolor{619D47}{RGB}{97,157,71}
\definecolor{DBDC4A}{RGB}{219,220,74}

\definecolor{D7191C}{RGB}{215,25,28}
\definecolor{FDAE61}{RGB}{253,174,97}
\definecolor{ABD9E9}{RGB}{171,217,233}
\definecolor{2C7BB6}{RGB}{44,123,182}

\pgfplotscreateplotcyclelist{diverge-4}{
    {D7191C!90!black,line width=1.5pt},
    {FDAE61!90!black,line width=1.5pt},
    {2C7BB6!90!black,line width=1.5pt},
    {ABD9E9!90!black,line width=1.5pt}
}

\pgfplotscreateplotcyclelist{diverge-3-rev}{
    {2C7BB6!90!black,line width=1.5pt},
    {FDAE61!90!black,line width=1.5pt},
    {D7191C!90!black,line width=1.5pt},
}

\definecolor{pastel1}{HTML}{fbb4ae}
\definecolor{pastel2}{HTML}{b3cde3}
\definecolor{pastel3}{HTML}{ccebc5}
\definecolor{pastel4}{HTML}{decbe4}
\definecolor{pastel5}{HTML}{fed9a6}
\pgfplotscreateplotcyclelist{pastel-1}{
    {pastel1!100!black,line width=1pt},
    {pastel2!100!black,line width=1pt},
    {pastel3!100!black,line width=1pt},
    {pastel4!100!black,line width=1pt},
    {pastel5!100!black,line width=1pt}
}

\definecolor{acc1}{HTML}{7fc97f}
\definecolor{acc2}{HTML}{beaed4}
\definecolor{acc3}{HTML}{fdc086}
\definecolor{acc4}{HTML}{ffff99}
\definecolor{acc5}{HTML}{386cb0}
\pgfplotscreateplotcyclelist{accent}{
    {acc1!100!black,line width=1pt},
    {acc2!100!black,line width=1pt},
    {acc3!100!black,line width=1pt},
    {acc4!100!black,line width=1pt},
    {acc5!100!black,line width=1pt}
}

\definecolor{dark1}{HTML}{1b9e77}
\definecolor{dark2}{HTML}{d95f02}
\definecolor{dark3}{HTML}{7570b3}
\definecolor{dark4}{HTML}{e7298a}
\definecolor{dark5}{HTML}{386cb0}
\pgfplotscreateplotcyclelist{dark}{
    {dark1!100!black,line width=1pt},
    {dark2!100!black,line width=1pt},
    {dark3!100!black,line width=1pt},
    {dark4!100!black,line width=1pt},
    {dark5!100!black,line width=1pt}
}

\definecolor{FEATURE}{HTML}{66c2a5}
\definecolor{CONTEXT}{HTML}{fc8d62}
\definecolor{REFINE}{HTML}{8da0cb}

\pgfplotscreateplotcyclelist{custompalette}{
    {3787CF!90!black,line width=1.5pt},
    {F1C232!90!black,line width=1.5pt},
    {619D47!90!black,line width=1.5pt},
    {DBDC4A!90!black,line width=1.5pt}
}

\pgfplotscreateplotcyclelist{custompalettethin}{
    {3787CF!90!black,line width=1.0pt},
    {F1C232!90!black,line width=1.0pt},
    {619D47!90!black,line width=1.0pt},
    {DBDC4A!90!black,line width=1.0pt}
}

\pgfdeclareplotmark{dot marker}{%
  \begin{pgfscope}%
    \color{black}%
    \pgfpathcircle{\pgfpointorigin}{\pgflinewidth / 4}
    \pgfusepath{fill}
  \end{pgfscope}%
}
\pgfplotsset{
  dot/.style={%
    mark=dot marker,%
    legend image code/.code={%
        \draw [line width=\pgflinewidth] (0cm, 0cm) -- (0.6cm, 0cm);%
    }%
  }
}

\pgfplotsset{every axis plot/.append style={line cap=round}}

\usepackage{listings}

\NewDocumentCommand{\headerot}{O{45} O{1em} m}{\makebox[#2][l]{\rotatebox{#1}{#3}}}%

\usepackage{stmaryrd}
\usepackage{trimclip}

\makeatletter
\DeclareRobustCommand{\shortto}{%
  \mathrel{\mathpalette\short@to\relax}%
}

\DeclareRobustCommand{\veryshortto}{%
  \mathrel{\mathpalette\veryshort@to\relax}%
}

\newcommand{\short@to}[2]{%
  \mkern2mu
  \clipbox{{.3\width} 0 0 0}{$\m@th#1\vphantom{+}{\shortrightarrow}$}%
  }

\newcommand{\veryshort@to}[2]{%
  \mkern2mu
  \clipbox{{.2\width} 0 0 0}{$\m@th#1\vphantom{+}{\shortrightarrow}$}%
  }

\def\convertto#1#2{\strip@pt\dimexpr #2*65536/\number\dimexpr 1#1}

\makeatother

\newcommand{\new}[1]{{#1}}

\newcommand*{\inparagraph}[1]{\medskip\noindent\textbf{#1}\hspace{0.5em}}

\usepackage{siunitx}
\usepackage{nicefrac}

\newcommand{\tablewidth}[0]{8.1cm}

\usetikzlibrary{positioning}
\usetikzlibrary{calc}

\newcommand{\oursS}[0]{ReCoVEr-MN}
\newcommand{\oursM}[0]{ReCoVEr-RN}
\newcommand{\oursL}[0]{ReCoVEr-CX}

\usepackage{etoolbox}           %
\newcommand{\B}[1]{\bfseries #1}

\usepackage[normalem]{ulem}

\robustify\uline
\def\U#1{#1\llap{\uline{\phantom{#1}}}}

\newcolumntype{R}[2]{%
    >{\adjustbox{angle=#1,lap=\width-(#2)}\bgroup}%
    c%
    <{\egroup}%
}
\newcommand*\rot{\multicolumn{1}{R{45}{1em}}}
\newcommand*\rottwo{\multicolumn{1}{R{45}{1.2cm}}}
\usepackage{bigdelim}
\usepackage{adjustbox}

\def\minDotSize{0pt}
\def\maxDotSize{16pt}
\def\minMemSize{0}
\def\maxMemSize{16}

\tikzset{declare function={size(\x)=((\maxDotSize-\minDotSize)*((\x-\minMemSize)/(\maxMemSize-\minMemSize)) + \minDotSize);}}

\tikzstyle{arrow} = [->,>=stealth]
\tikzset{
  object/.style={
    draw,
    rectangle,
    minimum width=2cm,
    minimum height=1cm,
    text centered,
    draw=black
  },
  arrow/.style={
    ->,
    >=Triangle
  },
  bridging path/.initial=arc,
  bridging span/.initial=8pt,
  bridging gap/.initial=2pt,
  bridge/.style 2 args={
    spath/split at intersections with={#1}{#2},
    spath/insert gaps after
    components={#1}{\pgfkeysvalueof{/tikz/bridging span}},
    spath/join components upright
    with={#1}{\pgfkeysvalueof{/tikz/bridging path}},
    spath/split at intersections with={#2}{#1},
    spath/insert gaps after
    components={#2}{\pgfkeysvalueof{/tikz/bridging gap}},
  }
}

\usepackage{placeins}

\usepackage[accsupp]{axessibility}  %

\definecolor{iccvblue}{rgb}{0.21,0.49,0.74}
\usepackage[pagebackref,breaklinks,colorlinks,allcolors=iccvblue]{hyperref}

\title{Removing Cost Volumes from Optical Flow Estimators}

\author{Simon Kiefhaber\textsuperscript{\normalfont{} 1,2} \qquad Stefan Roth\textsuperscript{\normalfont{} 1,2} \qquad Simone Schaub-Meyer\textsuperscript{\normalfont{} 1,2}\\
[0.5em]
\normalsize
\textsuperscript{1}Department of Computer Science, Technical University of Darmstadt \\ 
\normalsize
\textsuperscript{2}Hessian Center for AI (hessian.AI)\\
{\tt\small\url{https://visinf.github.io/recover}}
}

\usepackage{fancyhdr}
\usepackage{setspace}

\fancypagestyle{firststyle}
{

	\fancyhf{}
	\lfoot{{\footnotesize\begin{spacing}{.5}\parbox{\linewidth}{\vspace{-0.85em}%
					To appear in \emph{Proceedings of the IEEE/CVF International Conference on Computer Vision}, Honolulu, Hawai'i, USA, 2025.%
					\\\hrule\vspace{\baselineskip}
					\copyright~2025 IEEE. Personal use of this material is permitted. Permission from IEEE must be obtained for all other uses, in any current or future media, including reprinting/republishing this material for advertising or promotional purposes, creating new collective works, for resale or redistribution to servers or lists, or reuse of any copyrighted component of this work in other works. 
	}\end{spacing}}}
}

\begin{document}
    \twocolumn[{%
        \renewcommand\twocolumn[1][]{#1}%
        \maketitle%
        {\centering\includestandalone[width=0.99\textwidth]{figures/teaser}}%
        \captionof{figure}{\textbf{ReCoVEr.} We propose a method to remove cost volumes from optical flow estimators during training, and thereby, we are able to create fast and accurate optical flow estimators with a significantly reduced memory footprint. Our most accurate model, \oursL{}, reaches state-of-the-art accuracy while being more efficient \wrt inference and memory than SEA-RAFT \cite{Wang:2024:SEA}. Our most efficient model, \oursS{}, predicts sharper motion boundaries compared to the popular PWC-Net \cite{Sun:2018:PWC,Sun:2020:MMT}, while having comparable efficiency. \label{fig:teaser}\\[-1pt]}
    }]


\begin{abstract}
    Cost volumes are used in every modern optical flow estimator, but due to their computational and space complexity, they are often a limiting factor 
    regarding both processing speed and the resolution of input frames. Motivated by our empirical observation that cost volumes lose their importance once
    all other network parts of, \eg, a RAFT-based pipeline
    have been sufficiently trained, we introduce a training strategy that allows removing the cost volume from optical flow estimators throughout training. This leads to significantly improved inference speed and reduced memory requirements.
    Using our training strategy, we create three different models covering different compute budgets. Our most accurate model reaches state-of-the-art accuracy while being $1.2\times$ faster and having a $6\times$ lower memory footprint than comparable models; our fastest model is capable of processing Full HD frames at $20\,\mathrm{FPS}$ using only $500\,\mathrm{MB}$ of GPU memory.
\end{abstract}
\thispagestyle{firststyle} 
\vfill

\section{Introduction}
\label{sec:intro}
 
The task of optical flow estimation is to compute the apparent 2D motion between two consecutive frames for each pixel. Optical flow is a core part of many downstream tasks such as video inpainting~\cite{Zhou:2023:IPT,Gao:2020:FGV,Ke:2021:OVO,Xu:2019:DFV}, video frame interpolation~\cite{Niklaus:2020:SSV,Dong:2023:VFI,Zhang:2023:EMA,Sim:2021:EVF}, and object tracking~\cite{Yao:2023:FMO}.

Since optical flow essentially amounts to a 2D search problem~\cite{Xu:2021:HRO} for every pixel, it is very expensive to compute due to the quadratic nature of the problem. 
Recent methods for optical flow prediction are usually based on deep learning. However, they require a part of the network to specialize in computing similarities across time. This part is often realized using non-learnable layers that calculate the similarity of the pixels between frames. These layers are often referred to as `cost volumes,' `correlation layers,' or, in the context of transformer-based architectures, `cross-attention.' All these layer types have in common that they measure the similarity between pixels by calculating the cosine similarity, or a closely related measure, between every pixel of one input frame to candidate matching pixels of the other. They all share the problem of quadratic growth in computational and space complexity with input resolution. Due to these layers using very similar computations, we mainly refer to them as \emph{cost volumes} in this work. %

\begin{figure}[t]
    \centering
    \fontsize{8}{10}\selectfont

\begin{tikzpicture}
    \begin{axis}[
        xlabel=resolution,
        ylabel=TFLOPS,
        ymin=0,
        xmin=128,
        xmax=2432,
        scale only axis=true,
        width=0.4\textwidth,
        height=0.20\textwidth,
        legend style={at={(0.01,0.99)}, anchor=north west},
        set layers,
        axis line style={on layer=axis foreground},
        xticklabel={$\pgfmathprintnumber{\tick}^2$},
        cycle list name=diverge-4,
        grid style={line width=.1pt, draw=gray!10},
        major grid style={line width=.2pt,draw=gray!50},
        ymajorgrids
    ]

\addplot[name path=feature, color=FEATURE, mark=diamond, mark repeat=2, mark phase=2] table[] {
128  0.028788424224000003
256  0.115688453664
384  0.262304415264
512  0.471310177824
640  0.746449157664
768  1.092534318624
896  1.5154481720639998
1024  2.022142776864
1152  2.6206397394240004
1280  3.320030213664
1408  4.130474901024
1536  5.063204050464
1664  6.130517458464
1792  7.345784469024
1920  8.723443973664
2048  10.279004411424001
2176  12.029043768864002
2304  13.991209580064
2432  16.184218926624002
};

\addplot[name path=context, color=CONTEXT, mark=triangle, mark repeat=2, mark phase=2] table[] {
128  0.013352599072
256  0.053410379296
384  0.120173346336
512  0.21364150019199998
640  0.33381484086400004
768  0.48069336835200005
896  0.6542770826559999
1024  0.8545659837759999
1152  1.081560071712
1280  1.335259346464
1408  1.6156638080319998
1536  1.922773456416
1664  2.256588291616
1792  2.617108313632
1920  3.004333522464
2048  3.4182639181120003
2176  3.8588995005760003
2304  4.326240269856
2432  4.820286225952
};

\addplot[name path=refine, color=REFINE, mark=square, mark repeat=2, mark phase=2] table[] {
128  0.005277908512
256  0.021111617056000002
384  0.047501131296
512  0.08444645123199998
640  0.13194757686400002
768  0.19000450819200002
896  0.25861724521599994
1024  0.33778578793599995
1152  0.427510136352
1280  0.5277902904640001
1408  0.638626250272
1536  0.7600180157759999
1664  0.8919655869759999
1792  1.034468963872
1920  1.187528146464
2048  1.351143134752
2176  1.525313928736
2304  1.7100405284159998
2432  1.9053229337920001
};

\addplot[color=FEATURE!30] fill between [of = feature and context];
\addplot[pattern = north west lines, pattern color=gray!100] fill between [of = feature and context];

\addplot[CONTEXT!30] fill between [of = refine and context];

\path [name path=xaxis]
      (\pgfkeysvalueof{/pgfplots/xmin},0) --
      (\pgfkeysvalueof{/pgfplots/xmax},0);
      
\addplot[REFINE!30] fill between [of = xaxis and refine, soft clip={domain=128:2432}];

\legend {feature network, context network, refinement network}

\end{axis}
\end{tikzpicture}
    \vspace{-1.5em}
    \caption{\textbf{Analysis of the computational expense.} Cumulative number of floating point operations (FLOPS) required for a single optical flow prediction using SEA-RAFT-M~\cite{Wang:2024:SEA} for various input resolutions. The cost volume of SEA-RAFT is part of the \textit{feature network}. In this work, we demonstrate a method to remove the \textit{feature network} during training, thereby eliminating a major part of the required compute operations at inference time.}
    \label{fig:flop_breakdown_searaft}
    \vspace{-0.5em}
\end{figure}
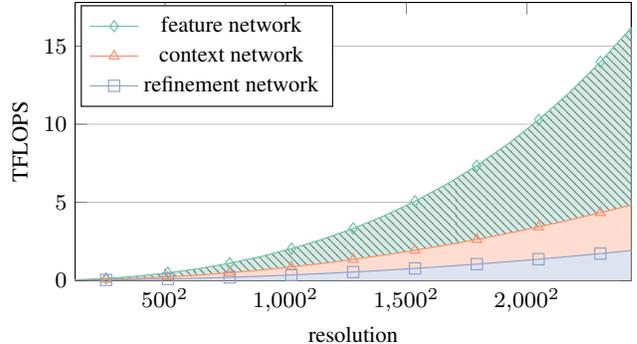

The recent SEA-RAFT approach~\cite{Wang:2024:SEA} achieves state-of-the-art accuracy while being relatively efficient. 
Still, the cost volume is responsible for more than half of the computations, as visualized in \cref{fig:flop_breakdown_searaft}, significantly dominating the computational cost. %
The cost volume also heavily influences and limits the maximum input resolution processable by current networks. The memory requirements of many methods increase so rapidly that even processing common Full HD ($1920 \times 1080$) frames can be problematic due to limited memory, as just the cost volume already requires $4\,\text{GB}$ at this resolution, growing to $62\,\text{GB}$ when the resolution is doubled.
Being able to remove or replace cost volumes is, therefore, a promising direction to significantly improve the efficiency of a wide range of current optical flow estimators.

Inspired by early research on this topic \cite{Dosovitskiy:2015:FN,Ilg:2017:FN2}, in this work, we analyze the role and benefit of cost volumes for the overall prediction error.
Based on our empirical observation that optical flow estimators are less dependent on their cost volume after training if they also have a context encoder for an initial prediction of the flow, we introduce a specific training strategy to adapt optical flow networks during training such that the cost volume is no longer needed at inference time. %
We propose three different models based on modified RAFT-like~\cite{Teed:2020:RAP} architectures, utilizing their two parallel branches: One branch computes features with a cost volume, while the other encodes context information and makes an initial optical flow prediction using an architecture without a cost volume (or similar). Together with our training strategy, we are able to remove the necessity of the cost volume completely during training. Our modified networks reach competitive accuracies, and our most powerful model even reaches state-of-the-art accuracies while being significantly faster in inference compared to previous models with comparable accuracies.

\section{Related Work}
\label{sec:rw}
\paragraph{Optical flow estimation.} 
Many different methods for optical flow estimation have been proposed over the years. Earlier methods often formulated optimization problems that were (approximately) solved to predict optical flow~\cite{Horn:1981:DOF,Lucas:1981:IIR,Black:1991:RDM,Papenberg:2006:HAO}. FlowNet~\cite{Dosovitskiy:2015:FN} was the first deep learning-based method that reached similar accuracies to classical methods by utilizing CNNs and correlation layers. Following the success of FlowNet, many different CNN-based methods like FlowNet2~\cite{Ilg:2017:FN2}, PWC-Net~\cite{Sun:2018:PWC,Sun:2020:MMT}, and SpyNet~\cite{Ranjan:2017:OFE} were proposed. All of these methods rely on the computation of some type of matching score in the form of either a correlation layer, cost volume, warping, or a combination of the aforementioned methods to determine the matching compatibility of features corresponding to pixels in both input frames. Early research on the FlowNet-S architecture \cite{Dosovitskiy:2015:FN} showed that CNNs without correlation layers are less accurate. We revisit this issue here.

Following the general trend of using vision transformers~\cite{Dosovitskiy:2021:IWW} in computer vision, many transformer-based methods like FlowFormer~\cite{Huang:2022:TAO}, CroCo-Flow~\cite{Weinzaepfel:2023:ICC}, and MemFlow~\cite{Dong:2024:OFE} utilize transformers. Still, notably, none of these approaches uses plain vision transformers, containing only self-attentions, but rather include some custom layers, like cost volumes or cross-attention. In contrast to other computer vision tasks, recent papers demonstrated that CNNs can still outperform ViTs for optical flow estimation while also being more efficient~\cite{Wang:2024:SEA,Dong:2023:ROF}.

\inparagraph{Cost volumes.} FlowNet~\cite{Dosovitskiy:2015:FN} introduced the concept of a correlation layer in the context of neural networks, where the similarity between a feature and its neighbors from one frame and all features of the other frame are computed. The computational complexity for calculating this layer is $\mathcal{O}(h^2w^2)$, where $h$ and $w$ refer to the height and width of the feature map, respectively. Since this layer is very expensive to compute at higher resolutions, FlowNet does limit the maximum displacement where the correlations are calculated to a fixed distance $D$, reducing the complexity to $\mathcal{O}(D^2hw)$. The disadvantage of this is that larger motions cannot be captured anymore.
To overcome the limited motion range to a certain degree, PWC-Net~\cite{Sun:2018:PWC,Sun:2020:MMT} and other methods~\cite{Hur:2019:IRR,Ranjan:2017:OFE,Hui:208:LCN} utilize image pyramids for a coarse-to-fine estimation where the flow is first estimated at a very low resolution and further upsampled and refined until the target resolution is reached. 
This allows to capture large motions at lower resolutions, where the corresponding pixel displacements are smaller, even when limiting the maximum displacements considered.
RAFT~\cite{Teed:2020:RAP} addresses the limitation of motion ranges by introducing a cost volume computed at multiple resolutions without any displacement range limitations. Instead of limiting the displacement range, RAFT limited the maximum resolution of the cost volume to $\nicefrac{1}{8}$ of the input frame resolutions, reducing the computational complexity enough to compute a global cost volume for the full feature map, though at a smaller feature resolution. The multi-scale cost volume is then sampled by a recurrent module that iteratively refines the flow prediction. The idea of globally matching the pixels was adapted and improved by multiple methods like GMA~\cite{Jiang:2021:LEH}, Flow1D~\cite{Xu:2021:HRO}, FlowFormer~\cite{Huang:2022:TAO}, CRAFT~\cite{Sui:2022:CAF}, and SEA-RAFT~\cite{Wang:2024:SEA}. The idea was also adopted by ViT-based~\cite{Dosovitskiy:2021:IWW} approaches, where cross-attention is used to compute similar features as a cost volume~\cite{Xu:2022:LOF, Xu:2023:UFS, Weinzaepfel:2023:ICC}.

\inparagraph{Efficient cost volumes.} Since cost volumes play a critical role in the overall accuracy of optical flow estimators while also using a significant amount of compute, multiple methods have been introduced to simplify the calculations in more sophisticated ways than the displacement range limitation used in earlier works. 
\citet{Jiang:2021:LOF} introduced the concept of sparse cost volumes, where a strategy is formulated to identify the top-$k$ best matching pixels from the other frame for each pixel and then only calculating the matching cost for these matches. Flow1D ~\cite{Xu:2021:HRO} introduced a decomposition of a cost volume into two lower-dimensional cost volumes and, therefore, approximated the solution of the 2D matching problem of optical flow by solving two 1D matching problems, allowing the calculation of optical flow between high-resolution input frames. As an alternative, HCVFlow~\cite{Zhao:2024:HCV} proposed the calculation of hybrid cost volumes that combine the top-$k$ matching idea of \citet{Jiang:2021:LOF} with the decomposition idea of Flow1D. 
\new{Instead of pre-calculating the entire cost volume, it is also possible to reduce the memory requirements of cost volumes by on-demand calculation of individual entries of the cost volume~\cite{Teed:2020:RAP, Jahedi:2024:HRM,Jahedi:2024:HRO}. While this approach reduces the memory footprint, it often increases the inference time significantly on commonly used accelerators. Recently, \citet{Briedis:2025:ECV} proposed to combine sparse evaluations of the cost volumes with specialized sampling strategies such that these calculations can be run more efficiently on common accelerators.}

\inparagraph{Efficient CNNs.} Not only the efficiency of cost volumes can be improved, but also the efficiency of CNNs. Over the years, many efficient architectures were proposed~\cite{Iandola:2016:ALA,Howard:2017:ECN,Sandler:2018:IRL,Howard:2019:SMN,Liu:2022:ACN,Xie:2017:ART,Chaudhuri:2019:RMS,Mehta:2018:ESP}. The efficiency improvements in CNNs are often reached by modifying kernel sizes~\cite{Xie:2017:ART, Liu:2022:ACN}, utilizing downsampling~\cite{Chaudhuri:2019:RMS,Liu:2022:ACN,Xie:2017:ART,Hesse:2023:CDC} or dilated convolutions~\cite{Mehta:2018:ESP}, and modified convolutional operators~\cite{Howard:2017:ECN,Sandler:2018:IRL,Howard:2019:SMN}.
We leverage this progress to realize efficient but powerful context networks for computing the initial optical flow. %

\section{Enhancing Optical Flow Estimators}
\label{sec:method}

Since the introduction of FlowNet~\cite{Dosovitskiy:2015:FN}
correlation or cost volumes are an integral part of many optical flow estimators. While in the original paper, the benefit of including these volumes over a pipeline without them was not yet really obvious, the subsequent work, FlowNet2~\cite{Ilg:2017:FN2}, clearly advocates for using them. However, cost volumes now heavily dominate the computational cost, including time and memory of an optical flow estimator, since their complexity is $\mathcal{O}(h^2w^2)$ for frames of size $\mathbf{I}\in\mathbb{R}^{3\times h \times w}$.
But since FlowNet, there have been multiple advancements in the building blocks of neural networks, and more powerful CNNs have been developed \cite{He:2016:DRL,Liu:2022:ACN}. This raises the question whether or in what form cost volumes are still needed 
or whether there are more efficient solutions with the same or even better accuracy.

\begin{figure}[t]
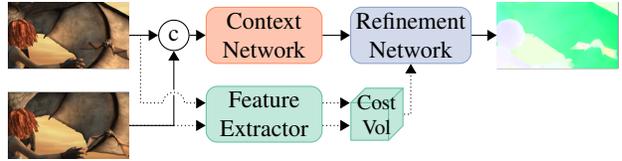

    \centering
    \includestandalone[]{figures/architecture}
    \vspace{-1em}
    \caption{\textbf{ReCoVEr architecture overview.} Our method assumes an architecture similar to RAFT~\cite{Teed:2020:RAP} where the input frames are processed by a \textit{context network} to obtain an initial flow estimate and context features, and in parallel, the inputs are processed by the \textit{feature network}, consisting of a feature extractor and a cost volume. The outputs of both branches are then combined by the \textit{refinement network} to obtain the optical flow prediction.
    Our training strategy allows us to cut away the \textit{feature network} during training (dotted path). This increases the computational and memory efficiency of the entire optical flow estimator at inference.}
    \label{fig:architecture}
    \vspace{-0.5em}
\end{figure}

\subsection{Analysis of SOTA optical flow pipelines}
We focus our analysis and proposed solution on the RAFT~\cite{Teed:2020:RAP} architecture, more precisely SEA-RAFT~\cite{Wang:2024:SEA}, as it represents the current state of the art \wrt endpoint error (EPE) while also being compute efficient. RAFT-based optical flow estimators mainly consist of three building blocks: a \textit{feature network}, a \textit{context network}, and a  \textit{refinement network}, as illustrated in \Cref{fig:architecture}. 
The feature network extracts features from each input frame and calculates a cost volume from these extracted features. In parallel to the feature network, the input is also processed by a context network, which usually only uses convolutional layers.
In the case of SEA-RAFT, the context network predicts an initial optical flow estimate as well as a feature map from the concatenated input frames. 
The outputs of the feature and context networks are then used as inputs for the refinement network, which iteratively refines the initial flow estimate by utilizing the cost volume and the additional features predicted by the context network to create the final optical flow prediction. 
While the feature network and the context network both produce information used for optical flow, they do not interact with each other directly and can individually be removed without breaking the network.
However, while the benefit of the context and refinement network has been ablated, the feature network with the cost volume has been treated as given. 
In the case of SEA-RAFT-M, the EPE on the Spring dataset \cite{Mehl:2023:HRH} improves by $0.62$ from the initial %
to the final prediction, but it remains unclear how big the contributions of the cost volume are to this result, in addition to just performing the refinement iterations.

We argue that the extra computation and memory needed for the cost volumes is disproportionally high \wrt to the rest of the framework. \Cref{fig:flop_breakdown_searaft} shows the FLOPS used for each component, and even for very efficient methods like SEA-RAFT, the computation of the cost volume takes $53\,\%$ of compute at low input resolutions like $480 \times 320$ and $61\,\%$ at higher resolutions like $1920 \times 1080$. Since the memory complexity grows at least as fast as the compute, this also leads to many methods being unable to even process Full HD ($1920\times 1080$) frames on a GPU with $48\text{GB}$ of VRAM.

Based on the high computational cost of cost volumes for an unclear impact on the accuracy, we conclude that having a closer look at the role and importance of the cost volume within the RAFT architecture is a promising direction towards significantly more efficient optical flow estimators.

\subsection{ReCoVEr}
\label{sec:removal}
Our goal is to \new{\textbf{Re}move \textbf{Co}st \textbf{V}olumes from optical flow \textbf{E}stimato\textbf{r}s (ReCoVEr)} while retaining the accuracy.

The naive approach, \ie having no cost volume and only using the context (here, a ResNet-based optical flow estimator) and refinement network (here, a convolutional GRU run for $4$ refinement iterations), indeed, does lead to significantly worse results (\cf \cref{tab:cost_fade}). Motivated by the observation of FlowNet2~\cite{Ilg:2017:FN2} that just modifying datasets and training schedules can lead to significant improvements, we analyze various strategies of reducing the contribution of the feature network during training. Specifically, we test \emph{(i)} the option of fading out the contribution of the cost volume by adding a dropout layer between the feature and refinement network with an increasing drop rate; \emph{(ii)} removing the entire feature network with the cost volume after a certain number of training steps (referred to as \textit{cut-off}). As we can see in \cref{tab:cost_fade}, both strategies lead to similar results but are significantly better than the refinement network that never had access to the cost volume. 
On the one hand, this empirically shows that the refinement module benefits from having access to the features from the cost volume when iteratively refining the flow, but on the other hand, this also hints that by just varying the training scheme, we can decrease the dependence of the refinement module on the cost volume for inference drastically.

However, the current EPE of this preliminary attempt is slightly worse than the original baseline, although being significantly faster ($2.5\times$) and more memory efficient ($8.8\times$). 
In the following, we further improve the training strategy and show that with the right choice of context network and training strategy, the EPE can be further reduced, leading to overall state-of-the-art results.

\begin{table}
  \centering

    \renewcommand{\arraystretch}{1.0}
    \setlength{\tabcolsep}{5pt}
    \small
    \begin{tabularx}{\linewidth}{@{}X c c c c c@{}}
        \toprule
          & \multicolumn{2}{c}{Sintel (val.)}  & \multirow{2}{*}[-0.5\dimexpr \aboverulesep + \belowrulesep + \cmidrulewidth]{Spring} & \multirow{2}{*}[-0.5\dimexpr \aboverulesep + \belowrulesep + \cmidrulewidth]{FLOPS} & \multirow{2}{*}[-0.5\dimexpr \aboverulesep + \belowrulesep + \cmidrulewidth]{memory}\\
         \cmidrule(lr){2-3}
         & Clean & Final & \\
        \midrule
        SEA-RAFT~\cite{Wang:2024:SEA} & \textbf{(0.43)} & \textbf{(0.58)} & \textbf{0.54} & \underline{4.40T} & \underline{8.21GB}\\
        \midrule
        no cost volume & 0.93 & 1.06 & 0.86 & \hspace{-0.8cm}\rdelim\}{3}{20pt}[] \\
        fade-out & 0.81 & 0.92 & \underline{0.68} & \textbf{1.74T} & \textbf{0.93GB}\\
        cut-off & \underline{0.80} & \underline{0.91} & \underline{0.68}\\
        \bottomrule
    \end{tabularx}

  \vspace{-0.5em}
  \caption{\textbf{Ablation of cost volume contributions}. Comparison between the endpoint errors (EPE) on the Sintel validation set \cite{Butler:2012:NOS} and the Spring training set \cite{Mehl:2023:HRH} when training a ResNet-based optical flow estimator without any cost volume compared to ResNet-based estimator where the cost volume is slowly faded out by increasing Dropout~\cite{Srivastava:2015:SWP} to $100\%$ over time, and one where the cost volume is cut off after a fixed number of training iterations. For completeness, we also show the accuracies achievable by an unmodified SEA-RAFT model. However, given that it was also trained on the Sintel validation set, we put these numbers in parentheses. The FLOPS and memory reported refer to \new{the fully trained networks in inference mode on} an input pair from the Spring dataset. \textbf{Bold} and \underline{underlined} values indicate the best and second best results.}
  \label{tab:cost_fade}
  \vspace{-0.5em}
\end{table}

\inparagraph{ReCoVEr training strategy.}
\label{sec:strategy}
To benefit from the cost volume during training but enable its removal during inference, we propose the following training strategy. Based on our analysis in \cref{tab:cost_fade}, we know that we can stop using the cost volume by fading out or completely cutting off the feature extraction branch after a certain number of training iterations, but not from the beginning. %
Therefore, we start training all parts of our optical flow estimator without any modifications to make sure that the training is stable and useful weights are learned for each part of our network. 
Since we have found no significant difference between fading out the cost volume compared to cutting it off at a fixed step, \cf \cref{tab:cost_fade}, we use the cut-off strategy and stop calculating the result of the feature network after a certain number of training iterations; we remove the parts of the refinement module that receive the cost volume as an input. Afterward, we continue training the shrunk-down network until it fully converges on the training dataset(s).

We mostly follow the training protocol of SEA-RAFT~\cite{Wang:2024:SEA}, where each model is trained in multiple stages. 
In the first stage, TartanAir~\cite{Wang:2020:DPL}, which consists of mostly rigid motions, is used, followed by training on the relatively simplistic motions and objects in FlyingChairs~\cite{Dosovitskiy:2015:FN}. Afterwards, the network is trained on FlyingThings~\cite{Mayer:2016:LDT}. For the final training stage, a combined dataset is created, denoted as TSKH, consisting of FlyingThings~\cite{Mayer:2016:LDT},  the FlowNet training split~\cite{Dosovitskiy:2015:FN} of Sintel~\cite{Butler:2012:NOS}, KITTI~\cite{Menze:2015:JEV,Menze:2018:OSF}, and HD1K~\cite{Kondermann:2016:BSS}. 
We decided to use this training protocol because it includes many different datasets, showcasing a wide variety of motions. Ablation studies in RAFT~\cite{Teed:2020:RAP} and SEA-RAFT have already shown that each part of this training stage improves the accuracies of their resulting models.

As this training strategy involves changing the datasets during training multiple times, a natural choice for the cut-off point is between swapping out the training datasets. \Cref{tab:cutoff} shows that by just varying the cut-off point, we can influence the EPE on Spring by $0.18$ pixels, and we conclude that it is best to remove the cost volume before the training on FlyingThings starts.

\begin{table}
  \centering

    \renewcommand{\arraystretch}{1.0}
    \setlength{\tabcolsep}{10pt}
    \small
    \begin{tabularx}{\linewidth}{@{}X c c c@{}}
        \toprule
        \multirow{2}{*}[-0.5\dimexpr \aboverulesep + \belowrulesep + \cmidrulewidth]{Cut-off} & \multicolumn{2}{c}{Sintel (val.)}  & \multirow{2}{*}[-0.5\dimexpr \aboverulesep + \belowrulesep + \cmidrulewidth]{Spring}\\
        \cmidrule(lr){2-3}
        & Clean & Final &\\
        \midrule
        Never & \textbf{(0.43)} & \textbf{(0.58)} & \textbf{0.54}\\
        \midrule
        TartanAir & 0.93 & 1.06 & 0.86\\
        FlyingChairs & \underline{0.80} & \underline{0.91} & 0.70\\
        FlyingThings & \underline{0.80} & \underline{0.91} & \underline{0.68}\\
        TSKH & 0.81 & 0.93 & 0.71 \\
        \bottomrule
    \end{tabularx}

  \vspace{-0.5em}
  \caption{\textbf{Analysis of training strategy.} Evaluation of the endpoint error (EPE) of a ResNet-based model after completing the entire training schedule. The cost volume is removed starting at the dataset mentioned in the ``cut-off'' column. The ``Never'' row denotes a model where the cost volume was never removed. Note that the training split of Sintel is part of the training, while Spring is not seen during training and, therefore, better shows the generalization ability of each model.}
  \label{tab:cutoff}
  \vspace{-0.5em}
\end{table}

\inparagraph{ReCoVEr backbones.} Until this point, we only evaluated the ResNet-$34$ context network proposed by SEA-RAFT, but in principle, every neural network that is capable of processing input frames and regressing dense features can be used as a context network. Currently, many optical flow methods utilize smaller ResNets as context networks~\cite{Teed:2020:RAP, Jiang:2021:LEH, Wang:2024:SEA, Jahedi:2022:CHC} because they offer a good trade-off between accuracy and compute complexity. %
Since we can remove the cost volume (during finetuning and inference), which is responsible for the largest amount of compute, this allows for the usage of more complex networks like Conv\-NeXt~\cite{Liu:2022:ACN} while still being faster than today's methods.

Specifically, we explore three different architectures for the context networks while keeping the feature and refinement network as proposed by SEA-RAFT.
All of our context networks take stacked input frames as input and return feature maps at $\nicefrac{1}{8}$ of the input resolution.
Our \emph{\oursM{}} model uses the first three residual blocks of a ResNet-$34$ as the context network. Thereby, the resulting network is equivalent to the SEA-RAFT-M architecture.
Our second model, \emph{\oursL{}}, is based on a ConvNeXt-t~\cite{Liu:2022:ACN} where we replace the last two downsampling-convolutions by convolutions of the same kernel size with stride $1$ to create feature maps at the required spatial dimensions.
The third model, \emph{\oursS{}}, utilizes a slightly modified version of the MobileNetV3-L~\cite{Howard:2019:SMN} encoder. We replaced the stride of the 13\,th block by $1$ to prevent the encoder from downsampling our features to $\nicefrac{1}{16}\,$th of the input resolution. %

As shown in \cref{tab:backbone}, our three proposed models offer different trade-offs: ConvNeXt is known to be very accurate, but expensive to compute~\cite{Liu:2022:ACN}, MobileNetV3 is a backbone with a very small memory footprint to optimize it for mobile devices, but compared to other CNNs its accuracy is limited~\cite{Howard:2019:SMN}, and ResNet generalizes very well across different computer vision tasks and usually offers a good trade-off between accuracy and computational efficiency~\cite{He:2016:DRL}.

\begin{table}
  \centering

    \renewcommand{\arraystretch}{1.0}
    \setlength{\tabcolsep}{5pt}
    \small
    \begin{tabularx}{\linewidth}{@{}X c c c c c@{}}
        \toprule
         \multirow{2}{*}[-0.5\dimexpr \aboverulesep + \belowrulesep + \cmidrulewidth]{Context network} & \multicolumn{2}{c}{Sintel (val.)}  & \multirow{2}{*}[-0.5\dimexpr \aboverulesep + \belowrulesep + \cmidrulewidth]{Spring} & \multirow{2}{*}[-0.5\dimexpr \aboverulesep + \belowrulesep + \cmidrulewidth]{FLOPS} & \multirow{2}{*}[-0.5\dimexpr \aboverulesep + \belowrulesep + \cmidrulewidth]{memory}\\
         \cmidrule(lr){2-3}
         & Clean & Final & \\
         \midrule
         MobileNetV3-L & 0.81 & \underline{0.90} & 0.99 & \textbf{0.86T} & \textbf{0.49GB}\\
         ResNet-34 & \underline{0.80} & 0.91 & \underline{0.68} & \underline{1.74T} & \underline{0.93GB}\\
         ConvNeXt-t & \textbf{0.36} & \textbf{0.42} & \textbf{0.51} & 2.65T & 1.24GB\\
         \bottomrule
    \end{tabularx}

  \vspace{-0.5em}  
  \caption{\textbf{Analysis of different backbones.} A comparison of EPEs reached by different context networks shows that the ConvNeXt-based network performs best on Sintel and Spring when all models are trained on the same data. The accuracies of ResNet and MobileNetV3 are almost identical on Sintel. The FLOPS and memory reported refer to an input frame resolution of $1920 \times 1080$.}
  \label{tab:backbone}
  \vspace{-0.5em}
\end{table}

\section{Results}
\label{sec:results}
To evaluate the performance of our models in more detail, we compare our networks against state-of-the-art methods like SEA-RAFT~\cite{Wang:2024:SEA} and more specialized versions of RAFT~\cite{Teed:2020:RAP}, like GMA~\cite{Jiang:2021:LEH} and GMFlow~\cite{Xu:2022:LOF}.

\inparagraph{Training details.}
We use the mixture of Laplace loss function introduced by SEA-RAFT for all of our trainings and a linear one-cycle learning rate scheduler~\cite{Smith:2017:VFT} with $6\text{k}$ iterations of warmup time and an AdamW optimizer~\cite{Loshchilov:2019:DWD}.
For a fair comparison, all of our trainings are done on $8$ NVIDIA RTX 6000 Ada (48 GB) GPUs for the exact same number of iterations, and we do not utilize any form of early stopping or model selection from the different states obtained during the training. We always report the accuracies of the models obtained after the last training step.

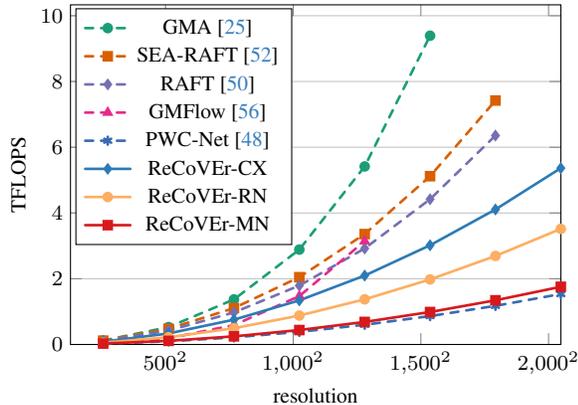
\begin{figure}
    \centering
    \fontsize{8}{10}\selectfont

\begin{tikzpicture}
    \begin{axis}[
        xlabel=resolution,
        ylabel=TFLOPS,
        ymin=0,
        xmin=128,
        xmax=2048,
        xticklabel={$\pgfmathprintnumber{\tick}^2$},
        legend style={at={(0.01,0.99)}, anchor=north west},
        width=\tablewidth,
        height=6.1cm,
        ytick={0, 2, 4, 6, 8, 10},
        cycle list name=dark,
        grid style={line width=.1pt, draw=gray!10},
        major grid style={line width=.2pt,draw=gray!50},
        ymajorgrids,
    ]

\addplot+[dashed, mark=*, mark options={solid, scale=0.75}] table [x=Resolution, y=TFLOPS, col sep=comma] {figures/data/complexity_GMA.csv};
\addlegendentry{GMA~\cite{Jiang:2021:LEH}}

\addplot+[dashed, mark=square*, mark options={solid, scale=0.75}] table [x=Resolution, y=TFLOPS, col sep=comma] {figures/data/complexity_SEARAFT.csv};
\addlegendentry{SEA-RAFT~\cite{Wang:2024:SEA}}

\addplot+[dashed, mark=diamond*, mark options={solid, scale=0.75}] table [x=Resolution, y=TFLOPS, col sep=comma] {figures/data/complexity_RAFT.csv};
\addlegendentry{RAFT~\cite{Teed:2020:RAP}}

\addplot+[dashed, mark=triangle, mark options={solid, scale=0.75}] table [x=Resolution, y=TFLOPS, col sep=comma] {figures/data/complexity_GMFlow.csv};
\addlegendentry{GMFlow~\cite{Xu:2022:LOF}}

\addplot+[dashed, mark=asterisk, mark options={solid, scale=0.75}] table [x=Resolution, y=TFLOPS, col sep=comma] {figures/data/complexity_PWCNet.csv};
\addlegendentry{PWC-Net~\cite{Sun:2020:MMT}}

\addplot+[2C7BB6, mark=diamond*, mark options={solid, scale=0.75}] table [x=Resolution, y=TFLOPS, col sep=comma] {figures/data/complexity_Ours_ConvNeXt.csv};
\addlegendentry{\oursL{}}

\addplot+[FDAE61, mark=*, mark options={solid, scale=0.75}] table [x=Resolution, y=TFLOPS, col sep=comma] {figures/data/complexity_Ours_ResNet.csv};
\addlegendentry{\oursM{}}

\addplot+[D7191C, mark=square*, mark options={solid, scale=0.75}] table [x=Resolution, y=TFLOPS, col sep=comma] {figures/data/complexity_Ours_MobileNetV3.csv};
\addlegendentry{\oursS{}}

\end{axis}
\end{tikzpicture}
    \vspace{-0.5em}
    \caption{\textbf{Comparison of the required number of floating point operations (FLOPS)} for representative optical flow estimators and our models at various resolutions. Missing data points are due to out-of-memory errors.
    }
    \label{fig:compare_flops}
    \vspace{-0.5em}
\end{figure}
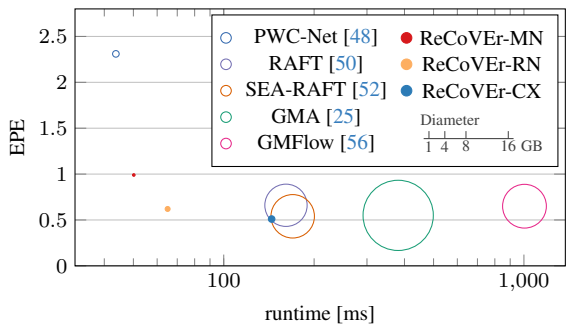
\begin{figure}[t]
    \centering

    \fontsize{8}{10}\selectfont
\begin{tikzpicture}[baseline=0pt]
\begin{axis}[
        width=\tablewidth,
        height=5cm,
        xlabel={runtime [ms]},
        ylabel={EPE},
        xmode=log,
        ytick={0, 0.5, 1.0, 1.5, 2.0, 2.5},
        log ticks with fixed point,
        scatter/classes={
            a={mark=o,dark1},
            b={mark=o,dark2},
            c={mark=o,dark3},
            d={mark=o,dark4},
            e={mark=o,dark5},
            x={mark=*,D7191C},
            y={mark=*,FDAE61},
            z={mark=*,2C7BB6}
        },
        ymin=0,
        ymax=2.8,
        grid style={line width=.1pt, draw=gray!10},
        major grid style={line width=.2pt,draw=gray!50},
        ymajorgrids,
        legend style={at={(0.98,0.98)}, anchor=north east, legend columns=1},
        legend style={font=\footnotesize},
        legend columns=2
    ]

\addplot[
        scatter, 
        only marks, 
        scatter src=explicit symbolic,
        visualization depends on = {size(\thisrow{Memory}) \as \perpointmarksize},
        scatter/@pre marker code/.append style={/tikz/mark size=\perpointmarksize},
    ] 
    table [x=Runtime, y=EPE, meta=label]{
    Model       EPE     Runtime Memory  label
    PWC-Net     2.31    43.72   1.25    e
    RAFT        0.66    161.19  8.00    c
    GMFlow      0.65    1004.23 8.28    d
    GMA         0.55    381.21  13.29   a
    SEA-RAFT    0.54    169.72  8.22    b
};
\addplot[
        scatter, 
        only marks, 
        scatter src=explicit symbolic, 
        visualization depends on = {size(\thisrow{Memory}) \as \perpointmarksize},
        scatter/@pre marker code/.append style={/tikz/mark size=\perpointmarksize},
    ] 
    table [x=Runtime, y=EPE, meta=label]{
    Model       EPE     Runtime Memory  label
    ReCoVEr-MN  0.99    50.16   0.49    x
    ReCoVEr-RN  0.62    65.03   0.93    y
    ReCoVEr-CX  0.51    144.41  1.24    z
};

\end{axis}

\pgfmathsetlengthmacro{\full}{2*\maxDotSize}

\node [draw, fill=white] at (4.12, 2.33) {
\begin{tikzpicture}
    \node [dark5] (pwc) at (0, 0) {\pgfuseplotmark{o}};
    \node [anchor=west, right of=pwc, node distance=1.2cm] (pwclabel) {PWC-Net~\cite{Sun:2020:MMT}};

    \node [dark3, below of=pwc, node distance=10pt] (raft) {\pgfuseplotmark{o}};
    \node [anchor=west, right of=raft, node distance=1.2cm] {RAFT~\cite{Teed:2020:RAP}};

    \node [dark2, below of=raft, node distance=10pt] (searaft) {\pgfuseplotmark{o}};
    \node [anchor=west, right of=searaft, node distance=1.2cm] {SEA-RAFT~\cite{Wang:2024:SEA}};

    \node [dark1, below of=searaft, node distance=10pt] (gma) {\pgfuseplotmark{o}};
    \node [anchor=west, right of=gma, node distance=1.2cm] (gmalabel) {GMA~\cite{Jiang:2021:LEH}};
    
    \node [dark4, below of=gma, node distance=10pt] (gmflow) {\pgfuseplotmark{o}};
    \node [anchor=west, right of=gmflow, node distance=1.2cm] {GMFlow~\cite{Xu:2022:LOF}};

    \node [D7191C, right of=pwclabel, node distance=1.2cm] (ours) {\pgfuseplotmark{*}};
    \node [anchor=west, right of=ours, node distance=1.0cm] {\oursS{}};
    
    \node [FDAE61, below of=ours, node distance=10pt] (ourm) {\pgfuseplotmark{*}};
    \node [anchor=west, right of=ourm, node distance=1.0cm] {\oursM{}};

    \node [2C7BB6, below of=ourm, node distance=10pt] (ourl) {\pgfuseplotmark{*}};
    \node [anchor=west, right of=ourl, node distance=1.0cm](llabel) {\oursL{}};

    \node [darkgray!100,inner sep=0, anchor=north, below of=llabel, node distance=16pt] {
    \begin{tikzpicture}
        \foreach \x/\y in {1pt/1,4pt/4,8pt/8,16pt/16} {
            \draw [line width=.2pt] (2*\x,0) -- (2*\x,0.1);
            \node [font=\fontsize{6pt}{0}\selectfont] at (2*\x,-0.1) {\y};
        }
        \draw (0,0) -- (34pt,0);
        \node [font=\fontsize{6pt}{0}\selectfont,anchor=west,inner sep=0] at (-0.05,0.25) {Diameter};
        \node [font=\fontsize{6pt}{0}\selectfont] at (41pt,-0.1) {GB};
    \end{tikzpicture}
    };
    
\end{tikzpicture}
};
            
\end{tikzpicture}
    \vspace{-0.5em}
    \caption{\textbf{Trade-offs} between runtime, memory usage, and accuracies for different models on Spring~\cite{Mehl:2023:HRH} in full resolution.}
    \label{fig:tradeoff}
    \vspace{-0.5em}
\end{figure}

\inparagraph{Evaluation details.}
Comparing the accuracies of existing optical flow methods in a comparable and fair setting
is tricky as training and evaluation datasets have changed over time.
Newer approaches, like SEA-RAFT, pre-train on the rigid motion of the TartanAir dataset before training on more traditional methods as this was shown to improve accuracies~\cite{Wang:2024:SEA}.
Further, datasets like Sintel~\cite{Butler:2012:NOS} and KITTI~\cite{Menze:2015:JEV,Menze:2018:OSF} do not have an official validation split, and consequently, the validation splits used by different methods differ, making a fair comparison on these datasets impossible since some methods have seen the evaluation data during training.
To overcome this, we evaluate all methods on the Spring dataset~\cite{Mehl:2023:HRH} without ever training any method on this dataset. The Spring dataset also has additional annotated regions in each frame with attributes like level of detail or rigidity of motions, allowing for further insights. We use the endpoint error (EPE)~\cite{Baker:2009:DBE,Otte:1994:OFE} for evaluating accuracy.

We use a batch size of $1$ and $\text{FP}32$ precision for measuring runtime and memory allocations. 
We average the runtime over $100$ iterations for all runtime measurements and perform $10$ warmup iterations that are not part of the measurements. The resolution reported in our figures is directly used as input sizes for all methods without applying additional down- and upsampling strategies for a fairer comparison of methods, as these kinds of strategies are not model-specific and could be applied for every model. We show the effects of using these strategies in the supplemental materials.

We compare our proposed models to GMA~\cite{Jiang:2021:LEH}, GMFlow~\cite{Xu:2022:LOF}, and SEA-RAFT~\cite{Wang:2024:SEA}, because all of them are extensions of RAFT where each method tackles a different shortcoming of RAFT. To clearly show the improvement of each of these methods over RAFT, we also included the original RAFT~\cite{Teed:2020:RAP} model. Additionally, we include PWC-Net~\cite{Sun:2018:PWC, Sun:2020:MMT} in our comparisons as it is very fast and memory efficient and, therefore, commonly  used in downstream tasks as a building block of other models. %

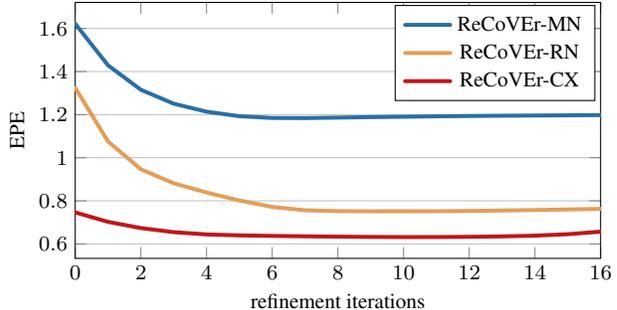
\begin{figure}
    \centering
    \fontsize{8}{10}\selectfont

\begin{tikzpicture}
    \begin{axis}[
        xlabel=refinement iterations,
        ylabel=EPE,
        xmin=0,
        xmax=16,
        legend style={at={(0.99,0.99)}, anchor=north east},
        cycle list name=diverge-3-rev,
        scale only axis=true,
        width=0.4\textwidth,
        height=0.195\textwidth,
        ytick={0.2, 0.4, 0.6, 0.8, 1.0, 1.2, 1.4, 1.6},
        grid style={line width=.1pt, draw=gray!10},
        major grid style={line width=.2pt,draw=gray!50},
        ymajorgrids,
    ]

\addplot table [x=Iterations, y=MobileNetV3, col sep=comma] {figures/data/iterations_cutoffT_spring.csv};
\addlegendentry{\oursS{}}

\addplot table [x=Iterations, y=ResNet, col sep=comma] {figures/data/iterations_cutoffT_spring.csv};
\addlegendentry{\oursM{}}

\addplot table [x=Iterations, y=ConvNeXt, col sep=comma] {figures/data/iterations_cutoffT_spring.csv};
\addlegendentry{\oursL{}}

\end{axis}
\end{tikzpicture}
    \vspace{-0.5em}
    \caption{\textbf{Evaluation of the effect of varying the number of refinement iterations} in our models on the resulting accuracy on every 20th training frame of Spring. In the case of $0$ iterations, the refinement network is not used, and the flow prediction of the context network is upsampled and used as a prediction.}
    \label{fig:iterations}
    \vspace{-0.5em}
\end{figure}

\begin{table}
    \centering
    \small
    \setlength{\tabcolsep}{5pt}
    \begin{tabularx}{\linewidth}{{@{}X c c c c@{}}} %
        \toprule
        \multirow{2}{*}[-0.5\dimexpr \aboverulesep + \belowrulesep + \cmidrulewidth]{Method} &  \multirow{2}{*}[-0.5\dimexpr \aboverulesep + \belowrulesep + \cmidrulewidth]{Middlebury~\cite{Baker:2011:DBE}} & \multicolumn{2}{c}{Monkaa~\cite{Mayer:2016:LDT}} & \multirow{2}{*}[-0.5\dimexpr \aboverulesep + \belowrulesep + \cmidrulewidth]{Spring~\cite{Mehl:2023:HRH}}\\
        \cmidrule(lr){3-4}
        & & Clean & Final & \\
        \midrule
        SEA-RAFT~\cite{Wang:2024:SEA} & \U{0.34}& \B{1.69} & \B{2.41} &  \U{0.54}\\
        \midrule
        \oursS{} & 0.60 & 2.88 & 3.03 & 0.99\\
        \oursM{} & 0.36 & 2.16 & 2.77 & 0.62\\
        \oursL{} & \B{0.33} & \U{1.71} & \U{2.46} & \B{0.51}\\
        \bottomrule
    \end{tabularx}
    \vspace{-0.5em}
    \caption{\textbf{Out-of-domain evaluations} of EPE on datasets that were not used during training without any finetuning.}
    \label{tab:ood_eval}
    \vspace{-0.5em}
\end{table}

\begin{table*}
  \centering
      \renewcommand{\arraystretch}{1.0}
    \setlength{\tabcolsep}{3pt}
    \small
    \begin{tabularx}{\linewidth}{@{}X c c | S@{\hspace{0.4cm}}  S S S S S S S S S S@{\hspace{-0.15cm}} S@{\hspace{0.6cm}}}
        \toprule
        Method & \rot{Sintel} & \rot{TartanAir} & \rot{total} & \rot{low-detail} & \rot{high-detail} & \rot{matched} & \rot{unmatched} & \rot{rigid} & \rot{non-rigid} & \rot{s0-10} & \rot{s10-40} & \rot{s40+}  & \rot{time[ms]} & \rottwo{memory[GB]} \\
        \midrule
        PWC-Net~\cite{Sun:2018:PWC,Sun:2020:MMT} & \crossmark & \crossmark & 2.31 & 2.27 & 10.84 & 2.13 & 11.28 & 2.10 & 4.45 & 1.83 & 3.16 & 15.87 & \B{43.72} & 1.25 \\
        RAFT~\cite{Teed:2020:RAP} & \checkmark & \crossmark & 0.66 & 0.61 & 9.66 & 0.54 & 6.55 & \U{0.24} & 4.91 & \U{0.18} & 2.85 & 10.83 & 161.19 & 8.00 \\
        GMFlow~\cite{Xu:2022:LOF} & \checkmark & \crossmark & 0.65 & 0.61 & \U{8.93} & 0.54 & \U{5.99} & 0.41 & \U{3.04} & 0.36 & 1.36 & \B{8.28} & 1004.23 & 8.28 \\
        GMA~\cite{Jiang:2021:LEH} & \checkmark & \crossmark & 0.55 & \U{0.50} & 11.35 & 0.45 & \B{5.97} & \B{0.20} & 4.13 & \B{0.16} & 1.82 & \U{10.33} & 381.21 & 13.29 \\
        SEA-RAFT~\cite{Wang:2024:SEA} & \checkmark & \checkmark & \U{0.54} & \U{0.50} & \B{8.76} & \U{0.40} & 7.25 & 0.31 & \B{2.90} & 0.20 & \B{0.95} & 10.56 & 169.72 & 8.22 \\
        \midrule
        \oursS{} & \checkmark & \checkmark & 0.99 & 0.93 & 14.36 & 0.82 & 9.48 & 0.54 & 5.57 & 0.28 & 1.95 & 21.96 & \U{50.16} & \B{0.49}\\
        \oursM{} & \checkmark & \checkmark & 0.62 & 0.59 & 9.48 & 0.49 & 7.77 & 0.34 & 3.60 & \U{0.18} & 1.24 & 13.78  & 65.03 & \U{0.93} \\
        \oursL{} & \checkmark & \checkmark & \B{0.51} & \B{0.47} & 
\U{9.10} & \B{0.39} & 6.51 & \U{0.24} & 3.28 & \B{0.16} & \U{1.08} & 10.80 & 144.41 & 1.24\\
        \bottomrule
    \end{tabularx}

  \vspace{-0.5em}
  \caption{\textbf{Comparison of the EPE} of different methods on the different annotated regions of the Spring training split. None of the methods were fine-tuned on Spring, and all of them were at least trained on FlyingChairs and FlyingThings, but since the training schedules changed over time, we marked the methods that were trained using additional data from TartanAir or Sintel. All evaluated methods received the full-resolution frames as inputs without any resizing. \new{All our models are faster (inference) and smaller than SEA-RAFT.}}
  \label{tab:spring_results}
  \vspace{-0.5em}
\end{table*}

\subsection{Quantitative results} 
\paragraph{Analysis of refinement.} Since the number of iterations in the refinement network is a hyperparameter, we also evaluate different numbers of iterations, as shown in \cref{fig:iterations}. The best accuracy for all of our models is reached between $7$ and $9$ iterations on our subset of Spring. Since the changes after $4$ iterations are minor for all of our models except for ResNet, we use only $4$ iterations for our MobileNetV3 and ConvNeXt-based models for all subsequent experiments and $8$ iterations for the ResNet-based model because the runtime tends to increase linearly with the number of refinement iterations~\cite{Wang:2024:SEA}.
The number of refinement iterations can be increased without retraining the model. %

\label{sec:complexity}
\inparagraph{Computational comparison. }
We compare the computational requirements of our models 
to other representative methods in \new{\cref{fig:compare_flops,fig:tradeoff}}. While the complexity of our three models differs quite a lot, they still require fewer operations than most other existing methods. Unsurprisingly, our \oursL{} model requires the most operations, followed by \oursM{} and \oursS{}. 
As can be seen, the number of computations for almost all existing methods tends to rise very quickly. This is due to the growing size of their cost volumes (\cf \cref{sec:complexity}), and since our models do not have any cost volumes after training, their growth in complexity is less steep, which allows for higher resolution images being processed.
The only model we evaluated that requires fewer computations than any of our models is PWC-Net. Its requirements are close to our fastest model, \oursS{}, but as we show in \cref{tab:spring_results} \new{and \cref{fig:tradeoff}}, our model is much more accurate at roughly the same
cost.
We observe similar trends for the inference time and memory requirements of the different models as for the number of floating-point operations. Further breakdowns of the different complexities %
can be found in the supplemental material.

\label{sec:generalization}
\new{
\inparagraph{Generalization capabilities.}
To quantify the generalization capabilities of our proposed methods, we evaluate SEA-RAFT and our methods on three different datasets that were not used during training. As shown in \cref{tab:ood_eval}, we find no significant differences between SEA-RAFT and our best method regarding the achieved accuracies. 
}

\inparagraph{Comparison to other methods.}
When comparing our models in \cref{tab:spring_results} to other representative methods that were proposed over the years, we find that our largest model, \oursL{}, is the best or second-best model for almost all accuracy measures, and overall best regarding total EPE. When comparing the accuracy of \oursL{} to SEA-RAFT for motions that are larger than $40$ pixels, we can see that our EPE is only $0.24$ higher for these regions, which equals less than $0.6\,\%$ of the entire motion magnitude. This shows that even though our networks do not have cost volumes anymore, modern convolutional networks are capable of having mostly the same accuracies for small and large motions, which differs from findings in the early works of FlowNet2~\cite{Ilg:2017:FN2} and FlowNet~\cite{Dosovitskiy:2015:FN}, showing the benefit of recent CNN innovations.
While \oursS{} and \oursM{} do not reach state-of-the-art performance, they are competitive with existing models while being faster. \Eg, \oursM{} has an accuracy that is very similar to that of RAFT, while being $2.5\times$ faster and using less than $\nicefrac{1}{8}$\,th of the memory. \oursS{} is almost as fast as PWC-Net but more than twice as accurate.
Generally speaking, \wrt to comparable computational budget, our ReCoVEr training strategy achieves much more accurate flow predictions than existing methods.

\begin{figure}
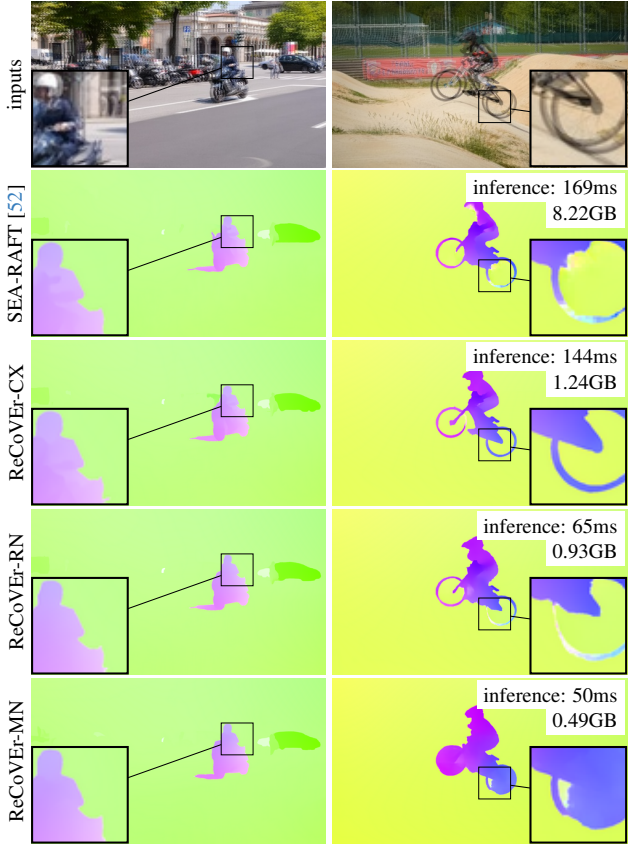

    \centering
    \includestandalone[width=\columnwidth]{figures/qual_sea_ours}
    \vspace{-1.5em}
    \caption{\textbf{Qualitative comparison} of our method compared to SEA-RAFT on Full HD frames from the DAVIS dataset~\cite{PontTuset:2017:DCV}. Note that our models use only between $\nicefrac{1}{6}$ to $\nicefrac{1}{20}$\,th of the memory required by SEA-RAFT and are $1.2$ to $3.4\times$ faster.}
    \vspace{-0.5em}
    \label{fig:qual_compare}
\end{figure}

\begin{figure}
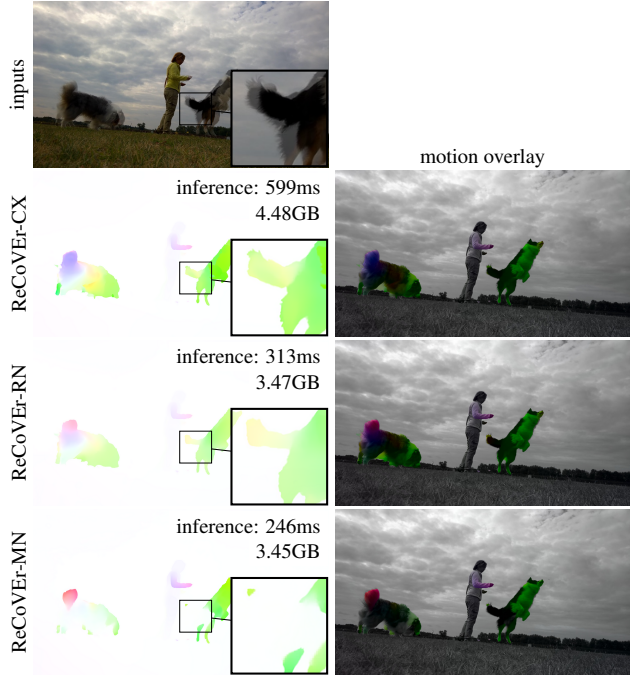

    \centering
    \includestandalone[width=\columnwidth]{figures/qual_4k}
    \vspace{-1.5em}
    \caption{\textbf{Qualitative result} for the prediction on 4K frames from the DAVIS dataset~\cite{PontTuset:2017:DCV}. Overlaying the prediction with the frames shows that the predictions are well-aligned with the input frames. The motion overlay shows a blend between the prediction and input frame where the color channel in LCh color space is taken from the optical flow visualization while the other channels are taken from the input frame.} 
    \label{fig:qual_4k}
    \vspace{-0.5em}
\end{figure}

\subsection{Qualitative results} 
Quantitatively evaluating optical flow on natural frames captured at high resolutions is impossible as no annotated, high-resolution dataset is currently available. In \cref{fig:qual_compare}, we therefore qualitatively compare the predictions of SEA-RAFT to the predictions of our models for two natural sequences from the DAVIS~\cite{PontTuset:2017:DCV} dataset captured at a resolution of $1920 \times 1080$. We find that both architectures using ResNet as a context network, \oursM{} and SEA-RAFT, have comparable prediction qualities and very similar failure cases where, \eg, the motion of the bike wheel is predicted very similarly.
In contrast, \oursL{} shows much sharper motion boundaries and is better at separating moving foreground objects from the background motion, especially visible at the wheels and the clear separation of the foreground and background motion, while still being faster and requiring less memory than SEA-RAFT.
The overall quality of \oursS{} prediction is noticeably lower than the others, but given the very low compute requirements, the overall quality of the results is still good, and when comparing it to the predictions of PWC-Net in \cref{fig:teaser}, \oursS{} shows an overall higher prediction quality at comparably low runtimes.

Even though our models are only trained on frames up to a resolution of $960\times 432$, we find that our models generalize to much higher resolutions, as can be seen in \cref{fig:qual_4k} where we predict the optical flow on an input at a resolution of $3840 \times 2160$ pixels. Similar to our findings for \cref{fig:qual_compare}, we observe the highest prediction quality for \oursL{}. \oursS{} shows some significant shortcomings when processing these high-resolution images, as can be seen by the holes in the prediction where no foreground motion is detected due to a lack of texture in that region.

\subsection{Limitations}\label{sec:limitations}
As shown in our quantitative and qualitative results, our fastest architecture, \oursS{}, starts failing for larger motions. The qualitative results show that \oursS{} in this case either fails to detect motions entirely or predicts motions going in the wrong direction. Further, the computational advantages of MobileNetV3 over ResNet start to diminish at high resolutions. Therefore, we conclude that the \oursS{} architecture should only be used for lower-resolution inputs in settings where a low memory footprint is the most important criterion. 
Further, we only explore the effect of replacing the context network and do not evaluate the effect other refinement network architectures can have on the performance. \Cref{fig:iterations} even shows that our refinement yields barely any increase in accuracy for \oursL{} and, therefore, removing the refinement module offers possibilities for further improvements.

\section{Conclusion}
\label{sec:conclusion}
In this work, we have analyzed and re-evaluated the role of cost volumes in optical flow estimators, taking into account the progress made in modern convolution network backbones and the availability of current datasets. We found that cost volumes are necessary initially during training, but with the right training strategy, they are not needed anymore during inference. %
To demonstrate optical flow methods without cost volumes, we introduced a simple yet highly effective training strategy where the cost volume is removed during the training process. We utilized this training strategy to create three different models that can cover a wide range of applications covering state-of-the-art accuracies, low compute times, and low memory footprints.

{\small \inparagraph{Acknowledgments.} 
This work was funded by the Hessian Ministry of Science and the Arts (HMWK) through the
project “The Third Wave of Artificial Intelligence -- 3AI”. 
The work was further supported by the Deutsche Forschungsgemeinschaft (German Research Foundation, DFG) – project number 529680848 and under Germany’s Excellence Strategy (EXC 3057/1 “Reasonable Artificial Intelligence”, Project No.\ 533677015). Stefan Roth acknowledges support by the European Research Council (ERC) under the European Union’s Horizon 2020 research and innovation programme (grant agreement No.\ 866008).}

{
    \small
    \bibliographystyle{ieeenat_fullname}
    \bibliography{bibtex/short,bibtex/papers,
    bibtex/local}
}

\clearpage

\maketitlesupplementary
\setcounter{page}{1}
\setcounter{table}{0}
\setcounter{figure}{0}
\setcounter{section}{0}

\renewcommand{\thepage}{\Roman{page}}
\renewcommand{\thefigure}{A.\arabic{figure}}
\renewcommand{\thetable}{A.\arabic{table}}
\renewcommand{\thesection}{A.\arabic{section}}

\section{Complexity}
In the main paper, we only refer to FLOPS as a measure of complexity. However, a reduction of FLOPS does not necessarily lead to a reduction in compute time on currently available accelerators, mostly due to memory alignment issues. Since we only remove entire parts of the networks and do not introduce sparsity or similar, we find that the reduction in FLOPS is proportional to the reduction in runtime. For completeness, we show the runtimes for different resolutions in \cref{fig:res_runtime}.

\begin{figure}
    \centering
    \fontsize{8}{10}\selectfont

\begin{tikzpicture}
    \begin{axis}[
        xlabel=resolution,
        ylabel={memory [GB]},
        ymin=0,
        xmin=256,
        xmax=2048,
        xticklabel={$\pgfmathprintnumber{\tick}^2$},
        legend style={at={(0.01,0.99)}, anchor=north west},
        cycle list name=diverge-4,
        grid style={line width=.1pt, draw=gray!10},
        major grid style={line width=.2pt,draw=gray!50},
        ymajorgrids,
        height=5cm,
        width=\tablewidth
    ]

\addplot table [x=Resolution, y=GB_peak, col sep=comma] {figures/data/complexity_Ours_ConvNeXt.csv};
\addlegendentry{\oursL{}}

\addplot table [x=Resolution, y=GB_peak, col sep=comma] {figures/data/complexity_Ours_ResNet.csv};
\addlegendentry{\oursM{}}

\addplot table [x=Resolution, y=GB_peak, col sep=comma] {figures/data/complexity_Ours_MobileNetV3.csv};
\addlegendentry{\oursS{}}

\addplot+[black, dotted] table [x=Resolution, y=GB_peak, col sep=comma] {figures/data/complexity_SEARAFT.csv};
\addlegendentry{SEA-RAFT~\cite{Wang:2024:SEA}}

\end{axis}
\end{tikzpicture}
    \vspace{-1.0em}
    \caption{\textbf{Memory} requirements of  our methods and SEA-RAFT for different input resolutions.}
    \label{fig:res_memory}
    \vspace{-0.5em}
\end{figure}
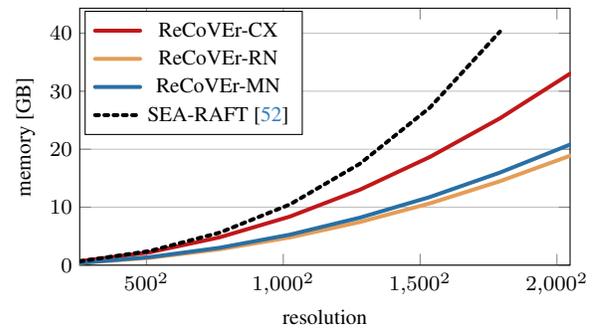

Another limiting factor is often the amount of memory required for a single prediction. We evaluate the memory footprint of our method in \cref{fig:res_memory}, and due to the missing cost volumes, we find a significant reduction in memory footprint. However, technically RAFT-style architectures never require every value of the cost volume to be available at the same time since the refinement network can only sample a certain number of values from the cost volume per iteration. Therefore, the required values could be calculated only when they are requested from the refinement module. This was also noticed and implemented by \citeauthor{Teed:2020:RAP} in the original implementation of RAFT, but the disadvantage of this approach is a significant slowdown in processing speed, and therefore, we do not consider this approach in our work.

\section{Downsample-upsample strategies}

\begin{table*}
  \centering
  \small
\setlength{\tabcolsep}{5pt}
\begin{tabularx}{\linewidth}{{@{}X c c c c c c c c@{}}} %
    \toprule
    \multirow{2}{*}[-0.5\dimexpr \aboverulesep + \belowrulesep + \cmidrulewidth]{Method} &  \multirow{2}{*}[-0.5\dimexpr \aboverulesep + \belowrulesep + \cmidrulewidth]{downsample} & \multicolumn{2}{c}{Sintel (val.)~\cite{Butler:2012:NOS}} & \multicolumn{2}{c}{Monkaa~\cite{Mayer:2016:LDT}} & \multirow{2}{*}[-0.5\dimexpr \aboverulesep + \belowrulesep + \cmidrulewidth]{Spring~\cite{Mehl:2023:HRH}} & \multirow{2}{*}[-0.5\dimexpr \aboverulesep + \belowrulesep + \cmidrulewidth]{time[ms]} & \multirow{2}{*}[-0.5\dimexpr \aboverulesep + \belowrulesep + \cmidrulewidth]{memory[GB]} \\
    \cmidrule(lr){3-4}
    \cmidrule(lr){5-6}
    & & Clean & Final & Clean & Final & \\
    \midrule
    \multirow{2}{*}{PWC-Net~\cite{Sun:2018:PWC,Sun:2020:MMT}} & \crossmark &  3.22 & 3.66 & 4.19 & 4.56 & 2.31 & 43.72 & 1.25\\
     & \checkmark & 5.27 & 5.95 & 5.16 & 5.34 & 3.83 & \U{12.30} & 0.36\\\rule{0pt}{10pt}%
    \multirow{2}{*}{RAFT~\cite{Teed:2020:RAP}} & \crossmark & (0.74) & (1.19) & 1.99 & 2.92 & 0.66 & 161.19 & 8.00 \\
     & \checkmark & (1.38) & (2.07) & 1.98 & 2.59 & 0.48 & 28.04 & 0.56\\\rule{0pt}{10pt}%
    \multirow{2}{*}{GMFlow~\cite{Xu:2022:LOF}} & \crossmark & (0.76) & (1.11) & 3.01 & 2.97 & 0.65 & 1004.23 & 8.28\\
     & \checkmark & (1.98) & (2.55) & 2.08 & 3.59 & 0.95 & 99.44 & 1.47\\\rule{0pt}{10pt}%
    \multirow{2}{*}{GMA~\cite{Jiang:2021:LEH}} & \crossmark & (0.62) & (1.06) & 1.75 & 2.63 & 0.55 & 381.21 & 13.29\\
     & \checkmark & (1.27) & (1.95) & 1.90 & 2.57 & \U{0.46} & 46.88 & 0.92\\\rule{0pt}{10pt}%
     \multirow{2}{*}{SEA-RAFT~\cite{Wang:2024:SEA}} & \crossmark & \U{(0.43)} & \U{(0.58)} & \B{1.69} & \U{2.41} & 0.54 & 169.72 & 8.22\\
    & \checkmark & (1.22) & (2.06) & 1.73 & \B{2.09} & \B{0.41} & 28.66 & 0.66\\
    \midrule
    \multirow{2}{*}{\oursS{}} & \crossmark & 0.81 & 0.90 & 2.88 & 3.03 & 0.99 & 50.16 & 0.49\\
     & \checkmark & 2.64 & 2.82 & 3.61 & 3.93 & 0.79 & \B{12.12} & \B{0.28} \\\rule{0pt}{10pt}%
    \multirow{2}{*}{\oursM{}} & \crossmark & 0.72 & 0.84 & 2.16 & 2.77 & 0.62 & 65.03 & 0.93\\
     & \checkmark & 1.85 & 2.33 & 2.52 & 3.05 & 0.57 & 13.46 & \U{0.30} \\\rule{0pt}{10pt}%
    \multirow{2}{*}{\oursL{}} & \crossmark & \B{0.36} & \B{0.42} & \U{1.71} & 2.46 & 0.51 & 144.41 & 1.24\\
     & \checkmark & 1.34 & 1.71 & 2.15 & 2.70 & 0.50 & 29.89 & 0.43 \\
    \bottomrule
\end{tabularx}

  \caption{\textbf{Effect of downsampling} by a factor of $2\times$ and bilinearly upsampling the resulting optical flow on different datasets. The time and memory refer to input frames of size $1920\times1080$. For completeness, we also show the accuracies achievable by methods that were also trained on the Sintel validation set, and we put these numbers in parentheses.}
  \label{tab:results_downsample}
\end{table*}

Downsampling the inputs and bilinearly upsampling the resulting optical flow is another method to reduce the memory footprint and inference time, and is, \eg, applied by SEA-RAFT on Full-HD frames to increase the computational efficiency for higher resolution inputs~\cite{Wang:2024:SEA}. In our work, we did not apply this orthogonal strategy as it can be applied theoretically to all optical flow methods. \Cref{tab:results_downsample} shows that when evaluating on the Spring dataset, most methods achieve even higher accuracies using the downsample-upsample strategy, but the results on Sintel and Monkaa clearly show that the increase in accuracy is not persistent between datasets.

\section{Additional technical details}
\paragraph{Refinement network.} The refinement network is implemented such that the sampling from the cost volume during the refinement can return an all-zero tensor instead of actually sampling from the cost volume. This simplifies the implementation of cutting away the feature network because by returning only zeros, the weights of the first layer of the refinement network that deal with this part of the input are not used since all of them are multiplied by zero. Theoretically, removing the affected weights from this layer completely would be possible. Still, this way of implementing the process is much easier, and the number of calculations needed to process the zero tensor is negligible compared to all other computations necessary for the optical flow prediction.

\inparagraph{Training protocol.} As described in \cref{sec:strategy}, we mostly follow the training protocol proposed by SEA-RAFT~\cite{Wang:2024:SEA}. The only difference is the composition of the TSKH dataset, where we do not include the validation split of Sintel. This allows us to fairly evaluate our methods on parts of the Sintel dataset for our analysis.

\begin{figure}
    \centering
    \fontsize{8}{10}\selectfont

\begin{tikzpicture}
    \begin{axis}[
        xlabel=resolution,
        ylabel={time [ms]},
        ymin=0,
        xmin=256,
        xmax=2048,
        xticklabel={$\pgfmathprintnumber{\tick}^2$},
        legend style={at={(0.01,0.99)}, anchor=north west},
        cycle list name=diverge-4,
        grid style={line width=.1pt, draw=gray!10},
        major grid style={line width=.2pt,draw=gray!50},
        ymajorgrids,
        height=5cm,
        width=\tablewidth
    ]

\addplot table [x=Resolution, y=ms, col sep=comma] {figures/data/complexity_Ours_ConvNeXt.csv};
\addlegendentry{\oursL{}}

\addplot table [x=Resolution, y=ms, col sep=comma] {figures/data/complexity_Ours_ResNet.csv};
\addlegendentry{\oursM{}}

\addplot table [x=Resolution, y=ms, col sep=comma] {figures/data/complexity_Ours_MobileNetV3.csv};
\addlegendentry{\oursS{}}

\addplot+[black, dotted] table [x=Resolution, y=ms, col sep=comma] {figures/data/complexity_SEARAFT.csv};
\addlegendentry{SEA-RAFT~\cite{Wang:2024:SEA}}

\end{axis}
\end{tikzpicture}
    \vspace{-1.0em}
    \caption{\textbf{Runtime} of our methods and SEA-RAFT for different input resolutions.}
    \label{fig:res_runtime}
    \vspace{-0.5em}
\end{figure}
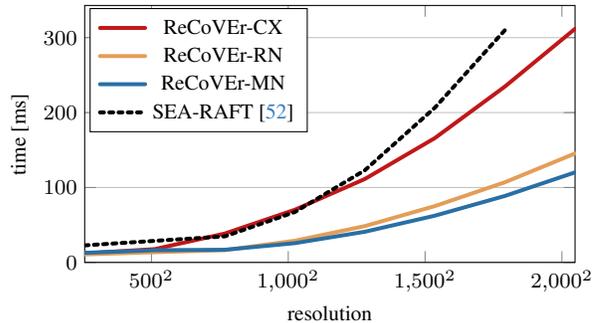

\section{Qualitative examples}
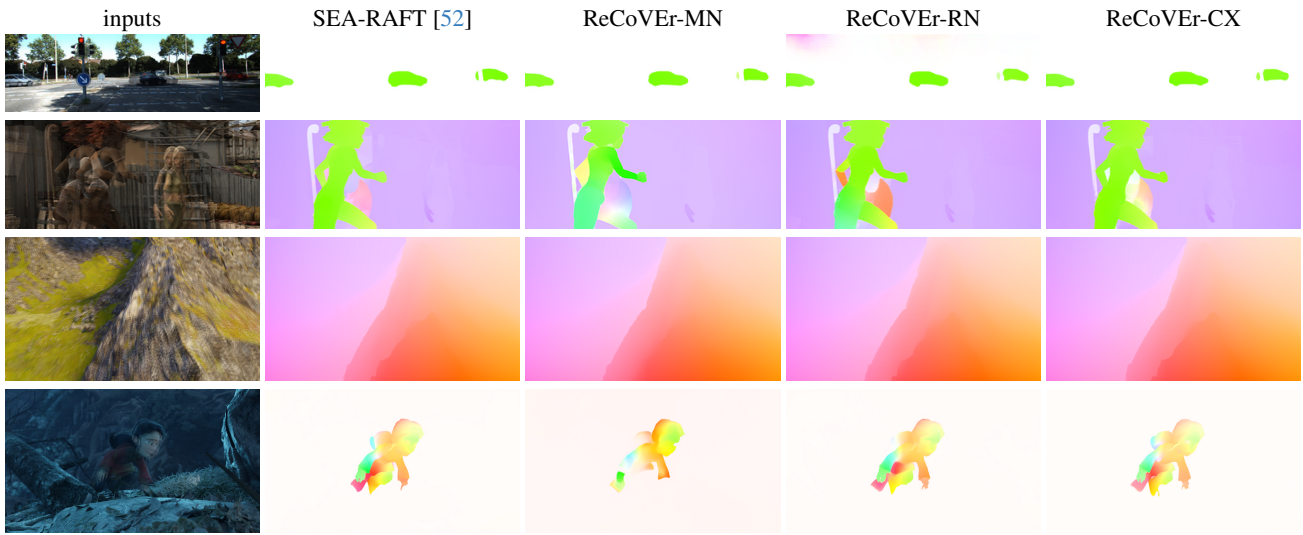
\begin{figure*}%

\fontsize{8}{10}\selectfont

\newcommand{\supplqual}[1]{\begin{tikzpicture}

    \node[inner sep=0pt] at (0, 0){
        \includegraphics[width=3cm]{figures/suppl/inputs/#1_1} 
    };
    \node[inner sep=0pt, opacity=0.5] at (0, 0){
        \includegraphics[width=3cm]{figures/suppl/inputs/#1_2} 
    };
\end{tikzpicture} & \includegraphics[width=3cm]{figures/suppl/sea/inputs-#1_1_inputs-#1_2_1} & \includegraphics[width=3cm]{figures/suppl/mobile/inputs-#1_1_inputs-#1_2_1} &  \includegraphics[width=3cm]{figures/suppl/resnet/inputs-#1_1_inputs-#1_2_1} & \includegraphics[width=3cm]{figures/suppl/convnext/inputs-#1_1_inputs-#1_2_1}}

\setlength{\tabcolsep}{1pt}

\begin{tabular}{c c c c c}
     inputs & SEA-RAFT~\cite{Wang:2024:SEA} & \oursS{} & \oursM{} & \oursL{} \\
    \supplqual{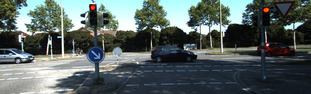}\\
    \supplqual{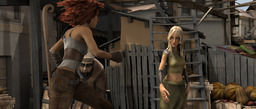}\\
    \supplqual{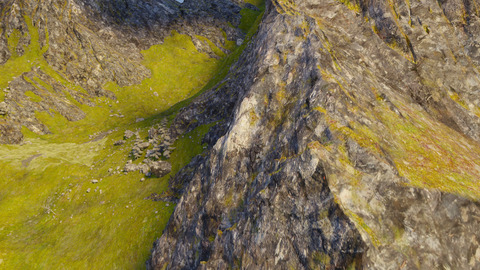}\\
    \supplqual{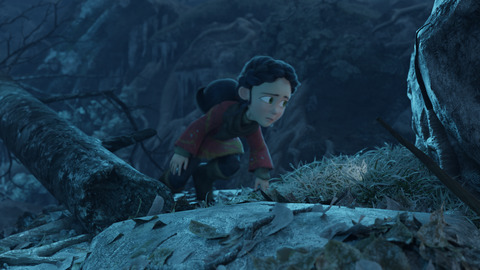}\\
\end{tabular}

    \caption{More qualitative examples on various frames taken from DAVIS~\cite{PontTuset:2017:DCV}, KITTI~\cite{Menze:2015:JEV,Menze:2018:OSF}, Sintel~\cite{Butler:2012:NOS}, and Spring~\cite{Mehl:2023:HRH}.}.
    \label{fig:suppl_qual2}
\end{figure*}

\begin{figure*}

\fontsize{8}{10}\selectfont

\newcommand{\supplqual}[1]{\begin{tikzpicture}

    \node[inner sep=0pt] at (0, 0){
        \includegraphics[width=3cm]{figures/suppl/inputs/#1_1} 
    };
    \node[inner sep=0pt, opacity=0.5] at (0, 0){
        \includegraphics[width=3cm]{figures/suppl/inputs/#1_2} 
    };
\end{tikzpicture} & \includegraphics[width=3cm]{figures/suppl/sea/inputs-#1_1_inputs-#1_2_1} & \includegraphics[width=3cm]{figures/suppl/mobile/inputs-#1_1_inputs-#1_2_1} &  \includegraphics[width=3cm]{figures/suppl/resnet/inputs-#1_1_inputs-#1_2_1} & \includegraphics[width=3cm]{figures/suppl/convnext/inputs-#1_1_inputs-#1_2_1}}

\setlength{\tabcolsep}{1pt}

\begin{tabular}{c c c c c}
     inputs & SEA-RAFT~\cite{Wang:2024:SEA} & \oursS{} & \oursM{} & \oursL{} \\
    \supplqual{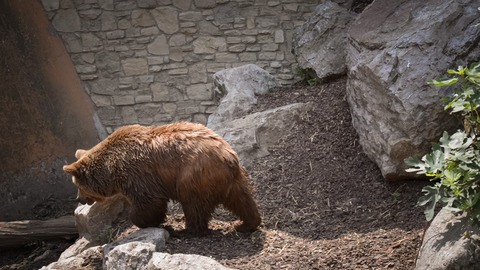}\\
    \supplqual{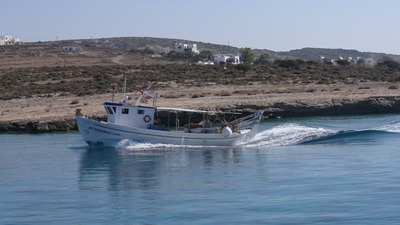}\\
    \supplqual{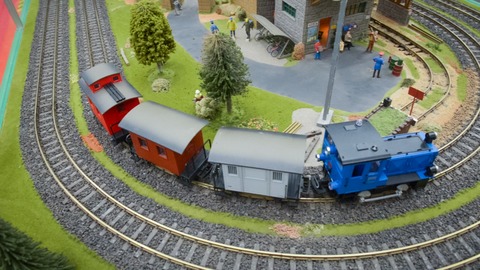}\\
    \supplqual{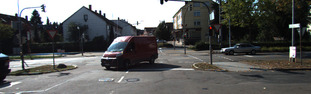}\\
    \supplqual{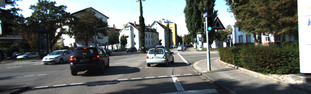}\\
    \supplqual{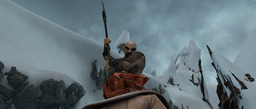}\\
    \supplqual{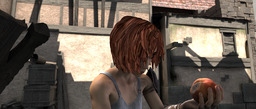}\\
    \supplqual{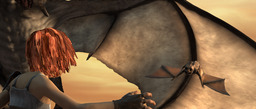}\\
    \supplqual{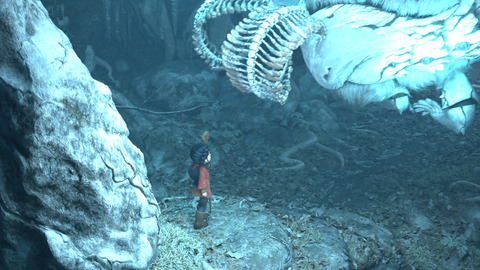}\\
    \supplqual{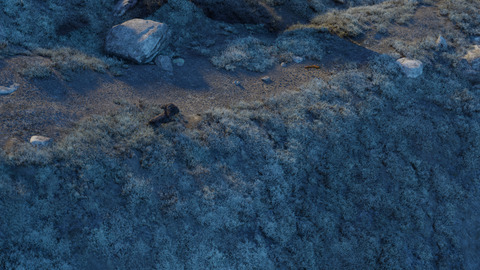}\\
    \supplqual{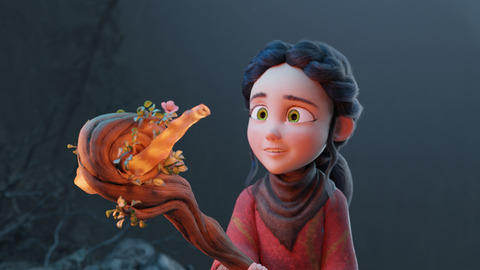}\\
\end{tabular}

    \caption{More qualitative examples on various frames taken from DAVIS~\cite{PontTuset:2017:DCV}, KITTI~\cite{Menze:2015:JEV,Menze:2018:OSF}, Sintel~\cite{Butler:2012:NOS}, and Spring~\cite{Mehl:2023:HRH}.}
    \label{fig:suppl_qual}
\end{figure*}

More qualitative examples, including samples from Sintel~\cite{Butler:2012:NOS}, Spring~\cite{Mehl:2023:HRH}, and natural images taken from KITTI~\cite{Menze:2015:JEV,Menze:2018:OSF} and DAVIS~\cite{PontTuset:2017:DCV} can be found in \cref{fig:suppl_qual,fig:suppl_qual2}.

\end{document}